\def\year{2020}\relax
\documentclass[letterpaper]{article} 
\usepackage{aaai20}  
\usepackage{times}  
\usepackage{helvet} 
\usepackage{courier}  
\usepackage[hyphens]{url}  
\usepackage{graphicx} 
\usepackage{graphicx}  
\usepackage{mathrsfs}
\usepackage{amsmath}
\usepackage{algorithm}
\usepackage[noend]{algpseudocode}
\usepackage{comment}
\usepackage{tablefootnote}

\DeclareMathOperator*{\argmax}{arg\,max}

\usepackage{amssymb}
\usepackage{color,xcolor,colortbl}
\usepackage{enumitem}
\usepackage{bm,bbm}
\usepackage{booktabs}
\usepackage{mathtools}
\usepackage{array}
\usepackage{subcaption}
\usepackage{pbox}
\usepackage{pifont}
\usepackage{multirow}
\usepackage{xcolor}

\newcommand\numberthis{\addtocounter{equation}{1}\tag{\theequation}}

\newcommand\like[1]{\begin{picture}(.8,.8)
\ifnum0=#1\put(.5,.25){\circle{.8}}\else
\ifnum10=#1\put(.5,.25){\circle*{.8}}\else
\put(.5,.25){\circle{.8}}\put(.5,.25){\circle*{.#1}}
\fi\fi\end{picture}}

\title{Controlling the Amount of Verbatim Copying in Abstractive Summarization}
\author{
Kaiqiang Song,$^\dagger$
Bingqing Wang,$^\ddagger$
Zhe Feng,$^\ddagger$
Liu Ren,$^\ddagger$ 
Fei Liu$^\dagger$\\
$^\dagger$Computer Science Department, University of Central Florida, Orlando, FL 32816, USA\\
$^\ddagger$Robert Bosch LLC, Sunnyvale, CA 94085, USA\\
kqsong@knights.ucf.edu, \{bingqing.wang, zhe.feng2, liu.ren\}@us.bosch.com, feiliu@cs.ucf.edu\\
}

\begin{document}

\maketitle

\begin{abstract}
An abstract must not change the meaning of the original text.
A single most effective way to achieve that is to increase the amount of copying while still allowing for text abstraction. 
Human editors can usually exercise control over copying, resulting in summaries that are more extractive than abstractive, or vice versa.
However, it remains poorly understood whether modern neural abstractive summarizers can provide the same flexibility, i.e., learning from single reference summaries to generate multiple summary hypotheses with varying degrees of copying.
In this paper, we present a neural summarization model that, by learning from single human abstracts, can produce a broad spectrum of summaries ranging from \emph{purely extractive} to \emph{highly generative} ones.
We frame the task of summarization as language modeling and exploit alternative mechanisms to generate summary hypotheses. 
Our method allows for control over copying during both training and decoding stages of a neural summarization model.
Through extensive experiments we illustrate the significance of our proposed method on controlling the amount of verbatim copying and achieve competitive results over strong baselines.
Our analysis further reveals interesting and unobvious facts.
\end{abstract}

\section{Introduction}
\label{sec:intro}

An ideal summarizer should provide the flexibility to generate summaries with varying proportions of reused text.
Such summaries are required to cater to diverse usage scenarios.
E.g., system abstracts may not contain excessive copied content without proper permission---11 consecutive words or longer are considered by EU standards as the author's intellectual creation and it is thus protected by copyright law~\cite{Castilho:2018}. 
Without proper control over copying, commercial summarizers can be held liable for copyright infringements. 
Moreover, system abstracts with an appropriate amount of copied content are more desirable than highly abstractive ones, as they are less likely to suffer from content hallucination~\cite{Reiter:2018:CL} and better at preserving the meaning of the original text.

To date, it remains poorly understood whether modern abstractive summmarization can provide the needed flexibility to control over copying and generate diverse abstracts.
Abstractive summarizers using encoder-decoder architectures can either copy words from the source text or generate new words unseen in the source~\cite{See:2017,Chen:2018:ACL,Gehrmann:2018}.
Recent work further attempted to increase the use of unseen words in summaries~\cite{Weber:2018,Kryscinski:2018}.
However, in all cases, the summarizers are trained on single-reference abstracts to produce single outputs with a fixed (corpus-level) copy rate. 
It can take multiple reference abstracts, created for the same input text with varying degrees of copying, to teach the system to generate abstracts with similar amounts of copying.
However, not only can it be time-consuming and costly to create human abstracts, but this is unlikely to be how humans learn to exercise control over copying.
Without an understanding of the copy mechanism of neural abstractive models, producing abstracts with varying degrees of copying can prove daunting at best and a ``mission impossible'' at worst.

\begin{table}[t]
\centering
\begin{small}
\begin{tabular}{l}
\toprule
\textbf{Question: What is the most probable next word?}\\
\textbf{Hint: the word is \textcolor{red}{seen} in the source text.}\\
\midrule
A 23-month-old toddler who was reportedly abducted in\\
Pennsylvania has been found dead, a district attorney said.\\
\midrule
Missing \_\_\_?\_\_\_\\
Missing Pennsylvania \_\_\_?\_\_\_\\
Missing Pennsylvania toddler \_\_\_?\_\_\_\\
Missing Pennsylvania toddler found \_\_\_?\_\_\_\\
\midrule
Reference Summary: \emph{Missing Pennsylvania toddler found dead}\\
\bottomrule
\toprule
\textbf{Question: What is the most probable next word?}\\
\textbf{Hint: the word is \textcolor{red}{unseen} in the source text.}\\
\midrule
Rescuers have suspended their search off the coast of Santa \\
Cruz Island for passengers who were trapped aboard the \\
Conception when the diving boat caught fire and sank.\\
\midrule
Search \_\_\_?\_\_\_\\
Search has \_\_\_?\_\_\_\\
Search has been suspended \_\_\_?\_\_\_\\
Search has been suspended in the \_\_\_?\_\_\_\\
Search has been suspended in the dive boat fire off \_\_\_?\_\_\_\\
\midrule
\textbf{Reference Summary}: \emph{Search has been suspended in the dive}\\
\emph{boat fire off California coast}\\
\bottomrule
\end{tabular}
\end{small}
\caption{Formulating summarization as a language modeling task.
The first model predicts only summary words that are \emph{seen} in the source text; the second model predicts only \emph{unseen} words.
Our method provides flexibility to control over copying by mix-and-matching the two types of behaviors.
}
\label{tab:example}
\end{table}

In this paper, our goal is to generate abstractive summaries with varying amounts of reused text by developing a general framework that learns from single reference summaries.
We define \emph{copy rate} as the percentage of summary $n$-grams appearing in the source text.
A high copy rate suggests that the summary is generated largely by copying verbatim from the source text.
Conversely, a low copy rate indicates there are more text shortening, word reordering, paraphrasing and abstraction involved in the generation process.
We argue that abstractive summarizers are not necessarily trained on \emph{every word} of reference summaries but they ought to separate the prediction of summary words that are \emph{seen} in the source text from those \emph{unseen}.
The underlying principle is simple and intuitively appealing. 
If a summarizer is trained to predict only \emph{seen} words, it learns to copy them from the source text, producing extractive summaries.
As more \emph{unseen} words are used for training, the summarizer gradually transforms from copying only to both copying and generating new words not present in the source text. 
By employing a ``mix-and-match'' strategy, we enable an abstractive summarizer to generate summaries with more, or less, copying.

We frame abstractive summarization as a language modeling task and present a \emph{decoder-only} framework for it.
It uses the same Transformer architecture~\cite{Vaswani:2017} to both encode the source text and decode the summary.
All network parameters are \emph{warm-started} using pretrained deep representations.
In contrast, in a typical encoder-decoder architecture, only parameters of the encoder and decoder can be warm-started but not those of the attention/copy mechanism~\cite{Khandelwal:2019}.
Further, our method allows for control over copying during both training and decoding stages of the neural model.
We experiment with varying proportions of seen and unseen summary words in training to teach the summarizer to favor, or not to favor, copying.
At decoding time, we compare different search strategies (best-first search vs. beam search) and reranking methods to encourage system abstracts to use wording similar to the original.
Despite that only single reference summaries are available in benchmark evaluations, we are able to evaluate summary quality along multiple dimensions, using automatic metrics based on lexical similarity (ROUGE; Lin, 2004)\nocite{Lin:2004} and semantic similarity (BERTScore; Zhang et al., 2019)\nocite{Zhang:2019:BERTScore}, and through human assessment of grammaticality, informativeness, and whether system abstracts remain true-to-original.
Our method demonstrates strong performance, either outperforming or performing on par with the best published results.
The research contributions are summarized as follows:
\begin{itemize}[topsep=5pt,itemsep=0pt,leftmargin=*]

\item we introduce a new summarization method that provides the needed flexibility to produce a spectrum of summaries for the same input and with a varying amount of copied content. 
Such summaries are highly desirable to cater to diverse real-world scenarios;\footnote{We make our implementation and models publicly available at\\ https://github.com/ucfnlp/control-over-copying}

\item our method emphasizes on in-depth analysis of the copy behavior in summarization. 
It frames abstractive summarization as a language modeling task and exploits multiple strategies at training and decoding stages to generate diverse summary hypotheses. 
We show competitive results and demonstrate the effectiveness of the proposed method on exercising control over copying. 

\end{itemize}

\section{Related Work}
\label{sec:related}

The significance of controlling over the copying behavior in summarization should not be underestimated.
Human editors often reuse the text in the original article to produce a summary~\cite{Jing:1999}.
But they can adjust the degree of copying to produce a wide spectrum of summaries.
E.g., human-written summaries for newswire~\cite{Over:2004,Hermann:2015}, meetings~\cite{Carletta:2005,Liu:2013:IEEETrans}, scientific articles~\cite{Qazvinian:2013} and online forums~\cite{Ouyang:2017} contain varying amounts of reused text. 
Moreover, the degree of copying can have a direct impact on scores of automatic evaluation metrics.
ROUGE was reported to favor summaries that use the same wording as the original~\cite{Ng:2015}.
If reference summaries are made by copying, system summaries with less copying and perhaps more abstraction, compression, and paraphrasing will be disadvantaged when compared against other system summaries with substantial copying.   
There is thus an urgent need, and this paper makes a first attempt to present a summarization framework that is capable of producing summaries with varying amounts of reused text.

To date, various extractive and abstractive summarization techniques have been investigated~\cite{Nenkova:2011}.
However, rarely has one technique been utilized to produce both extractive and abstractive summaries for any given text. 
Extractive summarization selects important and non-redundant sentences from the original document(s).
The sentences can be optionally compressed to remove inessential phrases, leading to compressive summaries~\cite{Martins:2009,Li:2013:EMNLP,Wang:2013,Filippova:2015,Durrett:2016}.
Abstractive summarization distills the source text into its essential meanings, then performs language generation from the representation to produce an abstract~\cite{Barzilay:2005,Liu:2015:NAACL,Liao:2018,Hardy:2018}.
These systems rarely provide the flexibility for an end user to indicate the desired amount of reused text in the summary.
To eliminate the need to develop multiple systems for extractive and abstractive summarization, 
we attempt to introduce control into the copying behavior of a neural abstractive summarization system.

Neural abstractive summarization has demonstrated considerable recent success.
It often utilizes an encoder-decoder architecture~\cite{Rush:2015,See:2017,Chen:2018:ACL,Lebanoff:2018,Celikyilmaz:2018}; and more recently, studies have attempted to use deep contextualized representations such as BERT~\cite{Devlin:2018} and ELMo~\cite{Peters:2018} to give a further boost to it. 
An encoder network converts the source text to a fix-length vector, conditioned on which a decoder network unrolls the summary one word at a time.
While it is tempting to use pretrained deep representations to ``warm-start'' the encoder/decoder, Khandelwal et al.~\shortcite{Khandelwal:2019} find that results can be less satisfying as the attention weights are still not pretrained.
In this paper we adopts a \emph{decoder-only} framework~\cite{Dong:2019} where the same Transformer architecture is used for both encoding the source text and decoding the summary.

Copying can help produce unseen words.
It was originally introduced to the seq2seq framework for neural machine translation~\cite{Gulcehre:2016} and later for abstractive summarization~\cite{See:2017}.
Particularly,
Knowles and Koehn~\shortcite{Knowles:2018} examine the influence of context and sub-words on the copying behavior of an NMT system.
To suppress copying, Kryściński et al.~\shortcite{Kryscinski:2018} introduce a novelty metric which is to be optimized during policy learning; and Weber et al.~\shortcite{Weber:2018} modify the scoring function of the summary sequence at decoding time. 
Fan, Grangier, and Auli~\shortcite{Fan:2018} attempt to control over summary length, entities, source style and portions.
But they do not address copying.
In this paper, we focus on better understanding the copying behavior of a summarization system and present effective mechanisms to control the amount of reused text.
We discuss what it takes for a summarizer to copy a word without an explicit copying mechanism, and how we may control the behavior to produce summaries with more, or less, copying.
In the following we describe our model in great detail.

\section{Our Approach}
\label{sec:our_approach}

We frame abstractive summarization as a language modeling task and present a \emph{decoder-only} framework for it.
It uses the same Transformer architecture~\cite{Vaswani:2017} to both encode the source text and decode the summary.
Let $\mathbf{x} = \{x_1, x_2, \dots, x_{|\mathbf{x}|}\}$, $x_i \in \mathcal{V}$ be a sequence of source tokens and $\mathbf{y}=\{y_1, y_2, \dots, y_{|\mathbf{y}|}\}$, $y_j \in \mathcal{V}$ be summary tokens.
Our goal is to model the conditional probability distribution $P(y_j|\mathbf{y}_{<j}, \mathbf{x})$ using a Transformer-inspired architecture.

\begin{figure}[t]
\centering
\includegraphics[width=3.2in]{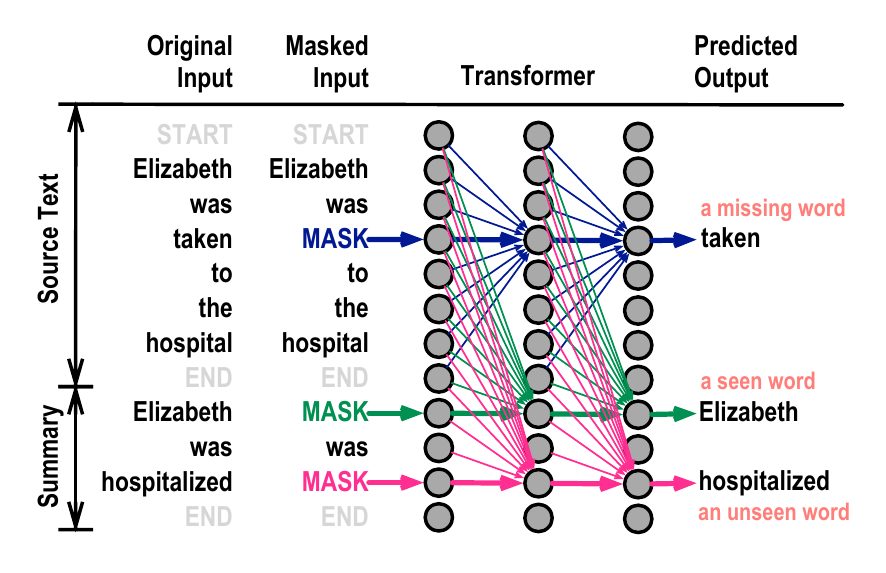}
\caption{An illustration of our CopyTrans architecture.
The self-attention mechanism allows 
(i) a source word to attend to lower-level representations of all source words (including itself) to build a higher-level representation for it, and 
(ii) a summary word to attend to all source words, summary words \emph{prior to} it, as well as the token at the current position (`MASK') to build a higher-level representation. 
}
\label{fig:architecture}
\end{figure}

We use byte-pair-encoding (BPE; Sennrich et al., 2016)\nocite{Sennrich:2016} for tokenization, with a vocabulary size of $|\mathcal{V}|=30,522$ tokens.
BPE has been shown to improve the robustness and accuracy of neural model training.
We use parameter tying, allowing the same token embeddings to be used in both the input layer and final softmax layer of the Transformer model.
Our method also includes three special tokens: \textsc{Start}, \textsc{End}, and \textsc{Mask}, which respectively denote the start/end of a sequence and a ``masked out'' token.
An illustration of our system architecture is provided in Figure~\ref{fig:architecture}.

\subsection{Training}
\label{sec:training}

We construct the source sequence $\mathbf{x}$ by prepending `\textsc{Start}' and appending `\textsc{End}' to the input text.  
E.g., $\mathbf{x}$ = \emph{START Elizabeth was taken to the hospital END}, illustrated in Figure~\ref{fig:architecture}.
Similarly, the target sequence $\mathbf{y}$ is constructed by appending `\textsc{End}' to the summary. 
E.g., $\mathbf{y}$ = \emph{Elizabeth was hospitalized END}.
Our system learns to predict the target sequence one word at a time until the `\textsc{End}' token has been reached. The conditional probability is shown in Eq.~(\ref{eq:model_y}-\ref{eq:model_z}).
\begin{align*}
P(\mathbf{y}|\mathbf{x})&=\prod_{j=1}^{|\mathbf{y}|}P(y_j | \mathbf{y}_{<j}, \mathbf{x})
\numberthis\label{eq:model_y}\\
&=\prod_{i=|\mathbf{x}|+1}^{|\mathbf{x}|+|\mathbf{y}|} P(z_i|\mathbf{z}_{<i})
\numberthis\label{eq:model_z}
\end{align*}

However, at training time, we argue that the system is not necessarily trained to predict \emph{every word} of target sequences but a selected collection might suffice. 
Using selected target tokens provides important potential to steer the system to be more extractive than abstractive, or vice versa.
We divide all tokens in the sequence $\mathbf{z} = [\mathbf{x};\mathbf{y}]$ into three categories: (a) summary tokens \emph{seen} in the source text, (b) summary tokens \emph{unseen} in the source, and (c) source tokens, with the expectation that training the system to predict only \emph{seen} summary tokens may reinforce the copying behavior, unseen tokens allow for generation, and source words enable the system to learn better token representations. 
By mix and matching target tokens from three categories, we enable a summarizer to generate summaries with more, or less, copying.

We randomly sample a set of tokens from each category using a Bernoulli distribution with probability $p$.
The value of $p$ varies by category and more analysis is provided in the experiments section.
Let $m_i \in \{0,1\}$ denote whether the $i$-th token of $\mathbf{z}$ is selected; its probability is defined as 
\begin{equation} 
\label{eq:Mask Rate}
P(m_i;p) = p^{m_i} (1-p)^{1-m_i}.
\end{equation}

A selected token is replaced by `\textsc{Mask}' 80\% of the time, meaning that the token has been `masked out' from the sequence $\mathbf{z}$.
For 10\% of the time, it is replaced by a random token from the vocabulary $\mathcal{V}$. It remains unchanged for the final 10\%.
In the following, we use $\mathbf{z}$ to represent the \emph{masked} sequence,
whose selected tokens are to be predicted during model training.
Our loss term is defined as follows:
\begin{align*}
\mathcal{L}(\theta)=-\sum_{i:m_i=1} \log\,P(z_i|\mathbf{z}_{\leq \max(i,|\mathbf{x}|)}).
\numberthis\label{eq:loss}
\end{align*}

It is important to note that we apply a binary mask to the self-attention mechanism of the Transformer architecture to allow 
(a) a \emph{source} token to attend to all source tokens including itself, and 
(b) a \emph{summary} token to attend to all source tokens, summary tokens \emph{prior to} it, as well as the current token (`\textsc{Mask}') in order to learn deep contextualized representations.
The formulation is similar to~\cite{Dong:2019}.
Our binary mask is defined by Eq.~(\ref{eq:AttentionMask}).
It is a square matrix whose $i$-th row represents the mask of the $i$-th token of $\mathbf{z}$.
If it is a source token ($i \leq |\mathbf{x}|$), the mask allows it to attend to all source tokens ($M_{i,j}^{\mbox{\scriptsize att}} = 1$ for $j \leq |\mathbf{x}|$).
If it is a summary token ($i > |\mathbf{x}|$), it can attend to all tokens prior to it as well as the current token ($M_{i,j}^{\mbox{\scriptsize att}} = 1$ for $j \leq i$).
\begin{equation} \label{eq:AttentionMask}
M_{i,j}^{\mbox{\scriptsize att}} = \left\{
\begin{split}
&1 \quad \text{if} \ j \leq \max(i, |\mathbf{x}|)\\
&0 \quad \text{otherwise}
\end{split}
\right.
\end{equation}

The input of Transformer consists of embedding matrices: $W_e$, $W_p$, and $W_s$ respectively denote the token, position, and segment embeddings~\cite{Devlin:2018}.
$\mathcal{Z}$, $\mathcal{P}$ and $\mathcal{S}$ are one-hot matrices used to retrieve embeddings for tokens in sequence $\mathbf{z}$.
The token, position, and segment embeddings for the $i$-th token are then added up element-wisely. 
\begin{align*}
E(\mathbf{z}) = \mathcal{Z}W_e + \mathcal{P}W_p + \mathcal{S}W_s
\numberthis\label{eq:Embedding}
\end{align*}

Our Transformer model takes as input embeddings $E(\mathbf{z})$ and the binary mask $M^{\mbox{\scriptsize att}}$ to produce a sequence of deep contextualized representations $\mathbf{h}=[\mathbf{h}_1, \mathbf{h}_2, \dots, \mathbf{h}_{|\mathbf{z}|}]$. 
Particularly, $\mathbf{h}_i$ is used to predict the $i$-th `missing' token in the sequence. 
We use parameter tying, allowing the same token embeddings $W_e$ to be used in both the input layer (Eq.~(\ref{eq:Embedding})) and final softmax layer of the model (Eq.~(\ref{eq:Prediction})).
\begin{align*}
& \mathbf{h} = \text{Transformer}(E(\mathbf{z}), M^{\mbox{\scriptsize att}})
\numberthis\label{eq:Transformer}\\
& P(z_i|\mathbf{z}_{\leq \max(i,|\mathbf{x}|)}) = \mbox{softmax}(W_e^\top \mathbf{h}_i)
\numberthis\label{eq:Prediction}
\end{align*}

\subsection{Decoding}
Given a trained model and an input text, the decoding stage searches for a summary sequence that maximizes $P(\mathbf{y}|\mathbf{x})$.
We present two search algorithms for this stage.

Best-first search uses a \emph{priority heap} to keep partial summaries, which are scored according to a heuristic function $f$. 
At each iteration, the search algorithm takes the highest-scoring partial summary, extends it by one word, then pushes new summary sequences back to the priority heap. 
We generate $k$ new summary sequences by selecting $k$ words that give the highest probability of $\log P(y_j|\mathbf{y}_{<j}, \mathbf{x})$ (Eq.~(\ref{eq:Searching})) then iteratively appending the words to the partial summary. 
If the highest-scoring summary in the heap concludes with an end-of-sentence symbol, it is moved to a pool of ``completed summaries'' for later reranking.
The heap thus keeps a collection of partial summaries of \emph{varying lengths}, which are visited according to their scores.\footnote{The size of the priority heap is capped at 1e5. If the heap has reached capacity and a new summary sequence needs to be pushed in, the lowest-scoring one will be removed from the heap.}
We provide an illustration of our best-first search algorithm in Algorithm~\ref{alg:AStar}.

In contrast, beam search is essentially breadth-first search.
It maintains a beam of size $k$ at any time step, containing partial summaries of the \emph{same length}.
For each partial summary, the algorithm extends it by one word, producing $k$ new sequences by appending each of the $k$ words that give the highest probability of $\log P(y_j|\mathbf{y}_{<j}, \mathbf{x})$ to the partial summary.
This process generates a total of $k*k$ new summary sequences by extending on each of the $k$ partial summaries.
The algorithm then selects k-best candidates, which are put in the beam for next iteration. 
If a candidate summary concludes with the end-of-sentence symbol, it is moved to the pool of ``completed summaries''.

Both best-first search and beam search employ the same scoring function that scores a candidate summary by the sum of log-likelihoods (Eq.~(\ref{eq:Searching})).
However, the two differ in their search strategies---beam search visits candidate summaries according to the summary length, whereas best-first search favors candidates attaining higher scores.
\begin{equation}
\label{eq:Searching}
\begin{split}
    &\mathbf{y}^{*}= \argmax_{\mathbf{y} \in \mathcal{Y}} \sum_{j=1}^{|\mathbf{y}|}\log P(y_j|\mathbf{y}_{<j}, \mathbf{x})\\
    &\mbox{s.t.} \quad y_{|\mathbf{y}|} = \textsc{End}
\end{split}
\end{equation}

\begin{algorithm}[t]
\caption{Best-First Search}\label{alg:AStar}
\begin{algorithmic}[1]
\Procedure{Best-First}{$src$, $\mathcal{M}, K$}
\newline
\Comment{Input sequence, model and beam size}
\State $\textit{init} \gets [\textsc{Start} || src|| \textsc{End}]$
\State $\mathcal{H}.push((\mathbf{0}, \textit{init}))$
\Comment{The priority queue}
\State $\mathcal{A}.reset()$
\Comment{The answer collector}
\While{($\mathcal{H}\textit{ is not Empty}) \textit{and} (\mathcal{A} \textit{ is not full})$}
    \State $\textit{current} \gets \mathcal{H}.pop()$
    \If{$\textit{current}$ ends with $\textsc{End}$}
        \State $\mathcal{A}.append(\textit{current})$
        \State $\textit{Continue}$
    \EndIf
    \State $\textit{Candidates}.reset()$
    \For{$\textit{each }\mathbf{w} \in  \mathcal{V}$}
        \State $\textit{extended} \gets \textit{current}\oplus \mathbf{c}$
        \State $\mathcal{S} \gets -log \mathcal{M}(\mathbf{w}|\textit{current}\oplus \textsc{Mask}) $
        \State $\textit{Candidates}.append((\mathcal{S}, \textit{extended}))$
    \EndFor
    \State $\textit{topK} \gets \textit{K-argmin }(\textit{Candidates})$
    \State $\mathcal{H}.pushAll(\textit{topK})$
\EndWhile
\Return $\mathcal{A}$
\EndProcedure
\end{algorithmic}
\end{algorithm} 
We compute $P(y_j|\mathbf{y}_{<j}, \mathbf{x})$ using our trained CopyTrans model.
Importantly, the `\textsc{Mask}' token is used as a prompt for the model to predict the next word.
E.g., ``\emph{START Elizabeth was taken to the hospital END Elizabeth was MASK}'' is a concatenation of the source text, partial summary and `\textsc{Mask}' token; it is fed to the CopyTrans model where the contextualized representation of `\textsc{Mask}' is used as input to a softmax layer to predict the next token $y_j \in \mathcal{V}$.
In experimental results, we demonstrate that a dynamic, contextualized representation of `\textsc{Mask}' performs reliably at predicting the next token. 
This represents an important distinction from shifting the target sequence by one position for prediction, which is common in encoder-decoder models.

\vspace{0.08in}
\noindent\textbf{Reranking}\quad
A reranking step is necessary, in part because candidate summaries decoded using beam search or best-first search do not always meet the length requirement. 
E.g., an overly short summary containing only two words is rarely an informative summary, despite that it may give a high log-likelihood score.
Below we compare three reranking strategies to offset this limitation. 

\textit{Length normalization} is adopted by See et al.~\shortcite{See:2017} and it is frequently used in many other systems. 
It divides the original log-likelihood score, denoted as $\mathcal{S}(\mathbf{x}, \mathbf{y}) = \log P(\mathbf{y}|\mathbf{x})$, by the total number of tokens in the summary to effectively prevent a long summary from being penalized.
\begin{align*}
\hat{\mathcal{S}}_{ln}(\mathbf{x},\mathbf{y}) &= \mathcal{S}(\mathbf{x},\mathbf{y})/|\mathbf{y}|
\numberthis\label{eq:s_bp}
\end{align*}

\textit{BP-norm} introduces a brevity penalty to summaries that do not to meet length expectation.
As illustrated in Eq.~(\ref{eq:s_bp}), BP-norm performs length normalization, then adds a penalty term $\log bp$ to the scoring function.
We modify the original penalty term of~\cite{yang-etal-2018-breaking} to make it favor summaries using more copying.
In Eq.~(\ref{eq:bp}), we define $r$ to be the copy rate, i.e., the percentage of summary tokens seen in the source text, scaled by a factor $c$.
When the copy rate $r$ is set to 1, the penalty is dropped to 0. 
Yang, Huang, and Ma~\shortcite{yang-etal-2018-breaking} provides a nice proof showing that this penalty term can directly translate to a coefficient multiplied to the log-likelihood score (Eq.~(\ref{eq:exp_s_bp})).
\begin{align*}
\hat{\mathcal{S}}_{bp}(\mathbf{x},\mathbf{y}) &= \log bp + \mathcal{S}(\mathbf{x},\mathbf{y})/|\mathbf{y}|
\numberthis\label{eq:s_bp}\\
bp &= \min(e^{1-1/r}, 1)
\numberthis\label{eq:bp}\\
\exp (\hat{\mathcal{S}}_{bp}(\mathbf{x},\mathbf{y})) &= bp \cdot \exp(\sum_{j=1}^{|\mathbf{y}|}\log P(y_j|\mathbf{y}_{<j}, \mathbf{x}))^{1/|\mathbf{y}|}\\
&= bp \cdot \Big[\prod_{j=1}^{|\mathbf{y}|} P(y_j|\mathbf{y}_{<j}, \mathbf{x})\Big]^{1/|\mathbf{y}|}
\numberthis\label{eq:exp_s_bp}
\end{align*}

\textit{Soft-bounded word reward (SBWR)} is a newly introduced method by us that assigns a per-word reward to the summary.
If the decoded summary is longer than expected ($i>\mathcal{L}_{\mbox{\tiny pred}}$),
the added words receive a diminishing reward of $\sigma(\mathcal{L}_{\mbox{\tiny pred}} - i)$. 
If the summary is shorter ($i \leq \mathcal{L}_{\mbox{\tiny pred}}$), every word of it will receive a reward.
The method thus promotes summaries of similar length to the predicted $\mathcal{L}_{\mbox{\tiny pred}}$.
A sigmoid function is used to smooth the reward values. $r$ is a coefficient to scale the total reward and it is tuned on the validation data.
\begin{align*}
\hat{\mathcal{S}}_{sbwr}(\mathbf{x},\mathbf{y}) = \mathcal{S}(\mathbf{x},\mathbf{y}) + r  \sum_{i=1}^{|\mathbf{y}|} \sigma(\mathcal{L}_{\mbox{\tiny pred}} - i)
\numberthis\label{eq:smooth-length-bounded-reward}
\end{align*}

We obtain the predicted length $\mathcal{L}_{\mbox{\tiny pred}}$ using greedy search, then empirically offset the predicted length by three words according to validation set.
In all cases, we force the decoder to never output the same trigram more than once during testing, which is a common practice to avoid repetitions~\cite{Paulus:2017}.

\begin{table}[t]
\centering
\setlength{\tabcolsep}{0pt}
\begin{small}
\begin{tabular}{l}
\toprule
\textbf{Source Text}: Premier Chang Chun-hsiung said Thursday he is \\
enraged and saddened by the snail-paced progress of the \\
reconstruction of areas hardest hit by a disastrous earthquake \\
that rattled Taiwan on Sept. 21 , 1999 .\\
\midrule
\textbf{Summary}:\\
1: premier expresses condolences for taiwan quake victims\\
2: premier angry over reconstruction of quake - hit areas\\
3: premier enraged and saddened by earthquake reconstruction\\
4: premier enraged by slow progress of post-quake reconstruction\\
\bottomrule
\toprule
\textbf{Source Text}: A blue-ribbon panel of experts said on Wednesday\\
that German economic growth will grind to a halt next year , \\
raising doubts about Berlin 's plans to shield Europe 's biggest \\
economy from the global turmoil .\\
\midrule
\textbf{Summary}:\\
1: german experts raise doubts about economic recovery\\
2: experts say german growth will grind to a halt next year\\
3: german experts to grind to halt next year\\
4: german economy will grind to halt in 2009 , say experts\\
\bottomrule
\end{tabular}
\end{small}
\caption{Example system summaries produced by
1: pointer-generator networks; 2: our method (best abstract), 3: our method (pure extract), and 4: human abstract.
}
\label{tab:output}
\end{table}

\begin{table*}[t]
\setlength{\tabcolsep}{3.8pt}
\renewcommand{\arraystretch}{1.08}
\centering
\begin{small}
\begin{tabular}{|ll|cccc|c|c|cccc|c|c|}
\hline
& & \multicolumn{6}{c|}{\textsc{Gigaword}} & \multicolumn{6}{c|}{\textsc{Newsroom}} \\[0.9mm]
& \textbf{Training Loss} & \textbf{1-gram} & \textbf{2-gram} & \textbf{3-gram} & \textbf{4-gram} & \textbf{Average} & \textbf{R-2}  & \textbf{1-gram} & \textbf{2-gram} & \textbf{3-gram} & \textbf{4-gram} & \textbf{Average} & \textbf{R-2} \\
\hline
a. & \like{8}\like{8}\like{8}\like{8} & 98.90 & 55.92 & 33.85 & 20.32 & 52.25 & 14.51 & 99.19 & 65.28 & 45.25 & 31.16 & 60.22 & 21.51\\
b. & \like{8}\like{8}\like{8}\like{8}\like{0} & 86.74 & 46.14 & 27.15 & 16.14 & 44.05 & 19.37 & 92.32 & 57.60 & 38.14 & 25.11 & 53.29 & \textbf{23.93}\\
c. & \like{8}\like{8}\like{8}\like{8}\like{0}\like{0} & 80.96 & 40.58 & 23.08 & 13.15 & 39.44 & \textbf{20.00} & 87.80 & 52.67 & 33.84 & 21.17 & 48.87 & 23.90\\
d. & \like{8}\like{8}\like{0}\like{0} & 73.98 & 34.89 & 19.19 & 10.55 & 34.65 & 19.20 & 82.23 & 46.55 & 28.54 & 17.37 & 43.67 & 22.97\\
\hline
e. & \like{8}\like{8}\like{8}\like{8}\like{2}\like{2} & 98.57 & 56.33 & 35.10 & 21.72 & 52.93 & 15.21 & 98.71 & 64.35 & 44.61 & 30.69 & 59.59 & 21.81\\
f. & \like{8}\like{8}\like{8}\like{8}\like{0}\like{2}\like{2} & 86.29 & 45.91 & 27.07 & 16.06 & 43.83 & 19.55 & 91.52 & 56.36 & 36.93 & 24.12 & 52.23 & \textbf{24.14}\\
g. & \like{8}\like{8}\like{8}\like{8}\like{0}\like{0}\like{2}\like{2} & 80.56 & 40.32 & 22.66 & 12.87 & 39.10 & \textbf{20.37} & 87.59 & 52.25 & 33.50 & 21.43 & 48.69 & 24.04\\
h. & \like{8}\like{8}\like{0}\like{0}\like{2}\like{2} & 74.22 & 35.09 & 19.13 & 10.49 & 34.73 & 19.39 & 82.41 & 47.16 & 29.16 & 17.92 & 44.16 & 23.10\\
\hline
\end{tabular}
\end{small}
\caption{The copy rate of various summarization models.
We define \emph{copy rate} as the percentage of summary $n$-grams appearing in the source text, where $n$=1/2/3/4 as well as an average of them.
We experiment with selecting varying amounts of \emph{seen} summary tokens (\like{8}\ ), \emph{unseen} summary tokens (\like{0}\ ), and \emph{source} tokens (\like{2}\ ) for training.
A circle corresponds to about 5 million tokens for \textsc{Gigaword} and 385k tokens for \textsc{Newsroom}, which are used to compute the loss term.
}
\label{tab:results_copy}
\end{table*}

\section{Experiments}
\label{sec:experiments}


\subsection{Data and Evaluation Metrics}
\label{sec:data}

We evaluate our proposed method on the sentence summarization task.
The goal is to condense a lengthy source sentence to a title-like summary.
Comparing to single-document summarization, sentence summarization deals less with content selection; 
its ground-truth summaries also contain more paraphrasing and abstraction. 
We conduct experiments on the Gigaword~\cite{Parker:2011} and Newsroom~\cite{Grusky:2018} datasets.
Gigaword articles were collected during 1995-2010 and Newsroom spans the range of 1998-2017.
We pair the first sentence of each article with its title to form an instance.
The train/valid/test splits contain 4 million/10k/1951 instances for Gigaword and 199k/21k/21k instances for Newsroom.
We experiment with both datasets to understand not only the copying behavior, but also domain adaptation effects for various models.
Despite that only single reference summaries are available in benchmark evaluations, we are able to evaluate summary quality along multiple dimensions, using automatic metrics based on lexical similarity (ROUGE; Lin, 2004)\nocite{Lin:2004} and semantic similarity (BERTScore; Zhang et al., 2019)\nocite{Zhang:2019:BERTScore}, and through human assessment of grammaticality, informativeness, and whether system abstracts remain true-to-original.

\subsection{Experimental Settings}
\label{sec:settings}

We initialize the model parameters using pretrained \textsc{Bert-Base} (uncased) model.
The model is fine-tuned on the training split of the Gigaword (or Newsroom) dataset for abstractive summarization.
Our model uses a 12-layer Transformer architecture. 
Its hidden state size is 768 and has 12 attention heads.
We use the Adam optimizer with $\beta_1=0.9, \beta_2=0.999$.
The learning rate is set to $lr$=4e-5 and it is halved whenever the validation loss does not change after 40,000 training steps.
We set the weight decay to be $0.01$ for regular layers and no weight decay for dropout and layer-normalization.
The sampling rate $p$ is set to 0.1 for source words and 0.9 for summary words, both seen and unseen. 
Each model is fine-tuned for 6 epochs; an epoch takes about 5 hours on a Tesla V100 GPU.
Our batch size is set to be 32.

\begin{table}[t]
\setlength{\tabcolsep}{5.5pt}
\renewcommand{\arraystretch}{1.08}
\centering
\begin{small}
\begin{tabular}{|l|ccc|c|}
\hline
\textbf{System} & \textbf{R-1} & \textbf{R-2} & \textbf{R-L} & \textsc{Bert-S}\\
\hline
lvt5k-1sent& 35.30 & {16.64} & 32.62 & -- \\
Multi-Task w/ Entailment & 32.75 & {15.35} & 30.82 & -- \\
SEASS & 36.15 & 17.54 & 33.63 & -- \\
DRGD & 36.27 & 17.57 & 33.62 & -- \\
EntailGen+QuesGen & 35.98 & 17.76 & 33.63 & -- \\
PG Networks & 34.19 & 16.92 & 31.81 & 58.32 \\
Struct+2Way+Relation & 35.61 & 18.02 & 33.45 & 58.84 \\
R3Sum & 36.36 & 18.23 & 33.85 & 56.74 \\
BiSET & 38.45 & 19.53 & 36.04 & 57.10 \\
\hline
\hline
Best-first Search & 39.07 & 20.28 & 36.49 & 61.27\\
Beam Search & 38.87 & 20.37 & 36.52 & 61.47 \\
Beam+LengthNorm &39.10 & 20.25 & 36.55 & 61.41\\
Beam+BPNorm (c=0.55) & \textbf{39.19} & 20.38 & \textbf{36.69} & 61.46 \\
Beam+SBWR (r=0.25) & 39.08 & \textbf{20.47} & 36.68 & \textbf{61.51} \\
\hline
\end{tabular}
\end{small}
\caption{Summarization results on the Gigaword test set.
The lower part of the table contains results from our system.}
\label{tab:results_gigaword}
\end{table}

\subsection{Summarization Results}
\label{sec:results}

\noindent\textbf{Control over copying}\quad
Could we bias a summarizer to produce summaries that are more extractive than abstractive, or vice versa?
If the summarizer is trained solely on summary words \emph{seen} in the source text, will it only learn to copy words during testing but not generate new words? 
We seek to answer these questions in this section.
Particularly, we divide all tokens selected for training into three categories: (a) summary tokens \emph{seen} in the source text, (b) summary tokens \emph{unseen} in the source, and (c) source tokens, with the expectation that training the system to predict only \emph{seen} summary tokens may reinforce the copying behavior, unseen tokens allow for generation, and source words enable the system to learn richer representations. 
By mix-and-matching tokens, we enable a summarizer to copy more, or less.

We analyze the copy rate of various summarization models in Table \ref{tab:results_copy}.
\emph{Copy rate} is defined as the percentage of summary $n$-grams appearing in the source text. 
We set $n$=1/2/3/4 and the average of them.
A high copy rate suggests that the summary is generated largely by copying verbatim from the source text.
We experiment with selecting varying amounts of \emph{seen} summary tokens (\like{8}\ ), \emph{unseen} summary tokens (\like{0}\ ), and \emph{source} tokens (\like{2}\ ) for training, where the number of circles is proportional to the number of tokens used in computing the loss term.
All summaries in Table \ref{tab:results_copy} are decoded using beam search (k=5) without reranking.

Our findings suggest that, the factor that makes the most impact on the copying behavior of a summarizer is the proportion of \emph{seen} and \emph{unseen} summary words used for training the model.  
If the summarizer is trained on purely \emph{seen} words (case a. in Table~\ref{tab:results_copy}), it only reuses source words during testing, despite that there is nothing to prevent the system from generating new words.
The 1-gram copy rate for case a. is about 99\% for both datasets, with the minor gap due to tokenization discrepancies.
As more \emph{unseen} words are used for training, the summarizer gradually transforms from copying only to both copying and generating new words not present in the source text. 
We observe that the ratio of seen vs. unseen words in ground-truth summaries is about 2:1 in both datasets, and \textsc{Newsroom} is slightly more extractive than \textsc{Gigaword}.
Our analysis reveals that it is important to maintain a similar ratio during training in order to achieve high ROUGE scores.
Pure extracts do not attain high ROUGE scores, as ground-truth summaries themselves are abstracts.
Our analysis further suggests that training on source words has little impact on the copying behavior of the system, but it improves representation learning and has lead to consistently improved ROUGE-2 F-scores.

\begin{table}[t]
\setlength{\tabcolsep}{5pt}
\renewcommand{\arraystretch}{1.08}
\centering
\begin{small}
\begin{tabular}{|l|l|ccc|c|}
\hline
& \textbf{System} & \textbf{R-1} & \textbf{R-2} & \textbf{R-L} & \textsc{Bert-S}\\
\hline
\multirow{4}{*}{\rotatebox{90}{\footnotesize Newsroom}} & PG Networks & 39.86 & 19.51 & 36.61 & 62.01\\
& Struct+2Way+Rel. & 40.54 & 20.44 & 37.40 & 62.05 \\
& Ours (pure-ext) & 43.21 & 21.81 & 40.05 & 63.68 \\
& Ours (best-abs) & \textbf{45.93} & \textbf{24.14} & \textbf{42.51} & \textbf{66.20}\\
\hline
\multirow{2}{*}{\rotatebox{90}{\footnotesize Giga}} & Ours (pure-ext) & 39.44 & 17.32 & 36.10 & 61.00 \\
& Ours (best-abs)& \textbf{40.89} & \textbf{19.11} & \textbf{37.60} & \textbf{62.74} \\
\hline
\end{tabular}
\end{small}
\caption{Summarization results on the Newsroom test set.
The top four systems are trained on Newsroom training data, whereas the bottom two systems are trained on Gigaword.}
\label{tab:results_newsroom}
\end{table}

\vspace{0.08in}
\noindent\textbf{System comparison}\quad
Table \ref{tab:results_gigaword} shows results on benchmark summarization data containing 1951 testing instances from Gigaword.
We contrast our system with summarization baselines developed in recent years.
They include {lvt5k-1sent}~\cite{Nallapati:2016},
{Multi-Task w/ Entailment}~\cite{Pasunuru:2018},
{SEASS} (Zhou et al., 2017),
{DRGD}~\cite{Li:2017:DRGD},
{EntailGen+QuesGen}~\cite{Guo:2018:ACL}, 
{PG Networks}~\cite{See:2017},
{Struct+2Way+Relation}~\cite{Song:2018},
{R3Sum}~\cite{Cao:2018:ACL}, and
{BiSET}~\cite{Wang:2019}.
Output summaries from the last four systems are graciously provided to us by the authors. 
We evaluate summary quality using two automatic metrics, including ROUGE\footnote{w/ options ``{\tt\scriptsize -c 95 -2 -1 -U -r 1000 -n 4 -w 1.2 -a -m}''} (Lin, 2004)\nocite{Lin:2004} that measures n-gram overlap between system and reference summaries, and BERTScore (Zhang et al., 2019)\nocite{Zhang:2019:BERTScore} that quantifies their semantic similarity using BERT-based contextualized representations.

Results show that our system achieves competitive performance, surpassing strong systems having reported results on this dataset, as judged by both metrics. 
These results demonstrate the effectiveness of our Transformer-based decoder-only architecture for abstractive summarization.
We observe that using beam search with reranking yields the highest results (using case g. in Table~\ref{tab:results_copy} for training).
Both BP-Norm and SBWR appear to be outstanding reranking methods, better than length normalization.
Our observation also suggests that best-first search and beam search can produce similar outcome, despite that the two differ in their search strategies, with beam search visiting candidates according to summary length and best-first search favoring candidates having high log-likelihood scores.
We suggest future work to explore other search methods such as A* search.

\vspace{0.08in}
\noindent\textbf{Domain adaptation}\quad
We investigate the effect of domain adaptation by training the model on Gigaword then testing it on Newsroom test set.
Results are reported in Table~\ref{tab:results_newsroom}.
Not surprisingly, there is a performance degradation when testing the model in a cross-domain setting.
We observe that the model with more copying (pure-extract, case e.) seem to degrade more gracefully than its counterpart (best-abstract, case f.), with a smaller performance gap in cross-domain settings.
Both of our models perform competitively comparing to other baseline methods.

\begin{table}[t]
\setlength{\tabcolsep}{5pt}
\renewcommand{\arraystretch}{1.08}
\centering
\begin{small}
\begin{tabular}{|l|cccc|}
\hline
\textbf{System} & \textbf{Inform.} & \textbf{Gramm.} & \textbf{Truthful.} & \textbf{Bst-Wst}\\
\hline
Human & 2.801 & 2.831 & 2.778 & -0.001\\
PG Networks & 2.768 & 2.697 & 2.678 & -0.058\\
R3Sum & 2.748 & 2.680 & 2.709 & -0.009\\
BiSET & 2.740 & 2.634 & 2.738 & -0.006\\
Ours (pure-ext) & \textbf{2.846} & 2.817 & 2.785 & 0.032\\
Ours (best-abs) & 2.843 & \textbf{2.855} & \textbf{2.865} & \textbf{0.042}\\
\hline
\end{tabular}
\end{small}
\caption{Human assessment of \emph{informativeness}, \emph{grammaticality}, \emph{truthfulness}, and best-worst scaling.}
\label{tab:human_giga}
\end{table}

\subsection{Human Evaluation}
To thoroughly analyze the quality of summaries, we ask human annotators to assess system outputs along three dimensions, including
\emph{informativeness} (Has the summary covered important content of the source text?), 
\emph{grammaticality} (Is the summary sentence grammatically correct?), 
and \emph{truthfulness} (Has the summary successfully preserved the meaning of the original text?).
Both system and human summaries are scored according to these criteria using a Likert scale from 1 (worst) to 5 (best).
We compare variants of our method generating (a) pure extracts (case e.) and (b) best abstracts (case g.), baselines of (c) PG networks, (d) R3Sum, (e) BiSET, and (f) human abstracts.
Following~\cite{liu-lapata-2019-hierarchical}, we perform Best-Worst Scaling
where a human selects the best and worst summary among six candidates.
The final rating of the system is computed as the percentage of times it was selected as the best minus that of the worst.
We sample 200 instances from the Gigaword test set for evaluation.
Each instance was assessed by five human evaluators from Amazon mechnical turk where low-quality annotations are manually removed. 
The results are presented in Table \ref{tab:human_giga}.
We observe that human summaries (article titles) are imperfect.
They can contain details that are nonexistent in the source (see Table~\ref{tab:output}), although they provide a means for researchers to train neural models without re-annotating reference summaries. 
In contrast, both of our systems perform slightly but consistently better than other baselines.

\section{Conclusion}
\label{sec:conclusion}

In this paper we present a Transformer-based, decoder-only framework to generate summaries with more, or less, copying.
The proposed method can be used to generate both extractive and abstractive summaries.
Our method emphasizes on in-depth analysis of the copy behavior in summarization. 
It exploits multiple strategies at training and decoding stages to generate diverse summary hypotheses. 
We show competitive results and demonstrate the effectiveness of the proposed method on exercising control over copying.

\section*{Acknowledgments}
We are grateful to the reviewers for their helpful comments. 
The work was performed in part while Kaiqiang Song was an intern at Bosch Research.
This research was supported in part by the National Science Foundation grant IIS-1909603.

\begin{small}
\bibliographystyle{aaai.bst}
\bibliography{reference,summ_aaai,abs_sum}
\end{small}

\end{document}